# Single-Nodal Spontaneous Symmetry Breaking in NLP Models

Shalom Rosner[b,1], Ronit D. Gross[a,1], Ella Koresh[a,1], and Ido Kanter[a,b,*]

[a]Department of Physics, Bar-Ilan University, Ramat-Gan, 52900, Israel.
[b] Gonda Interdisciplinary Brain Research Center, Bar-Ilan University, Ramat-Gan, 52900, Israel.

*Corresponding author at: Department of Physics, Bar-Ilan University, Ramat-Gan, 52900, Israel.
E-mail address: ido.kanter@biu.ac.il (I. Kanter).
[1]These authors equally contributed to this work

## Abstract

Spontaneous symmetry breaking in statistical mechanics primarily occurs during phase transitions at the thermodynamic limit where the Hamiltonian preserves inversion symmetry, yet the low-temperature free energy exhibits reduced symmetry. Herein, we demonstrate the emergence of spontaneous symmetry breaking in natural language processing (NLP) models during both pre-training and fine-tuning, even under deterministic dynamics and within a finite training architecture. This phenomenon occurs at the level of individual attention heads and is scaled-down to its small subset of nodes and also valid at a single-nodal level, where nodes acquire the capacity to learn a limited set of tokens after pre-training or labels after fine-tuning for a specific classification task. As the number of nodes increases, a crossover in learning ability occurs, governed by the tradeoff between a decrease following random-guess among increased possible outputs, and enhancement following nodal cooperation, which exceeds the sum of individual nodal capabilities. In contrast to spin-glass systems, where a microscopic state of frozen spins cannot be directly linked to the free-energy minimization goal, each nodal function in this framework contributes explicitly to the global network task and can be upper-bounded using convex hull analysis. Results are demonstrated using BERT-6 architecture pre-trained on Wikipedia dataset and fine-tuned on the FewRel classification task.

## 1. Introduction

One of the most fundamental phenomena in physics is spontaneous symmetry breaking which emerges in many different physical systems [1-4]. In dynamic systems, spontaneous symmetry breaking indicates that the low-temperature free energy exhibits reduced symmetry compared to that of the underlying theoretical description. It occurs without a noticeable macroscopic external field, and is typically induced by non-extensive preferential uniform or random external fields, or initial conditions that break the symmetry. An example from statistical physics is the ferromagnetic Ising system in two or more dimensions [5, 6], where at low temperatures, the system develops a macroscopic magnetization that spontaneously aligns either up or down, with both states possessing the same free energy. In such ordered systems, the time average of a microscopic part of the system, such as that of a single Ising spin, can reveal the macroscopic state of the entire system. Conversely, for disordered systems, such as a spin-glass system [7], the equilibrium state of the entire system cannot be deduced from the time-dependent frozen behavior of an individual spin.

Spontaneous symmetry breaking has recently been determined as a key mechanism underlying deep learning models [8, 9]. The deep learning architecture, which comprises disordered weights, splits the learning task among its parallel learning components comprising filters in convolutional neural networks [9, 10] and attention heads in transformer architectures [11-14]. Enhanced cooperation among these building blocks along the architecture's layers and within each layer enhances the signal-to-noise ratio and learning capabilities. A necessary component for the emergence of spontaneous symmetry breaking in deep networks is randomness in the initial conditions. If all filters or attention heads within a layer are initialized with identical parameters, weights and biases, symmetry among them is preserved, preventing spontaneous symmetry breaking, and neutralizing efficient learning.

This study demonstrates that spontaneous symmetry breaking in natural language processing (NLP) models can be scaled-down to the level of single nodes within attention heads during pre-training and fine-tuning NLP tasks [13-15]. This spontaneous symmetry breaking can be observed even at the single-node level, where a node is capable of learning a small number of tokens after the pre-training process or identifying several labels after fine-tuning for a specific classification task. Furthermore, the learning capability of several nodes exceeds the sum of their individual nodal learning capabilities as a result of cooperation. The results are demonstrated using BERT-6 architecture [16] trained using a small subset of the Wikipedia dataset [17] and fine-tuned using the fully supervised FewRel classification task [18]. These findings indicate that spontaneous symmetry breaking can occur even at the smallest scale of a finite network, which can evolve under deterministic dynamics without stochastic updates, similar to zero-temperature dynamics in infinite physical systems.

## 2. BERT-6 architecture

The behavior of each node in the attention head output and its role in facilitating learning for NLP tasks are presented using the BERT-6 architecture [19, 20], comprising the following three main parts (Fig. 1). The first is an embedding layer comprising 30,522 tokens representing each word or sub-word in the Wikipedia dataset (Fig. 1, left). Each token is embedded [21-23] into a 768 dimensional vector using an embedding layer. The embedding layer encodes the lexical meaning of a token and its position within the input. The length of the input sequence was fixed at 128 tokens, and any unused space is padded using a [PAD] token [24]. The second part of the architecture consists of six-transformer encoders comprising query, key, and value (QKV) attention, followed by projection and feedforward (FF) layers (Fig. 1, middle) [16]. The number of input and output dimensions for this part of the architecture is 768 per input token (Fig. 1, right). The QKV attention comprises 12 heads with 768 input and 64 output dimensions per head and input token (Fig. 1, right). The last part is a classifier head comprising two fully connected (FC) layers terminating at 30,522 outputs representing the tokens (Fig. 1, left, purple).

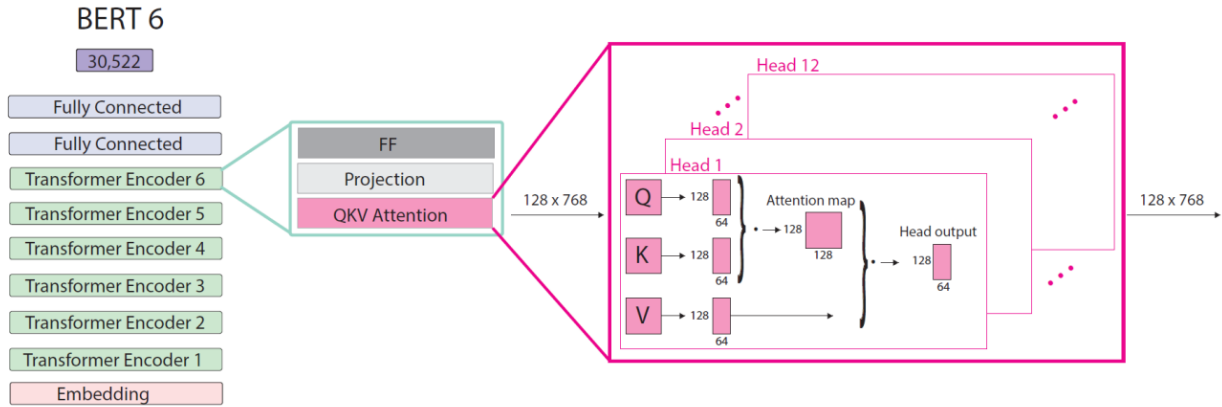

**Fig. 1.** Scheme of the BERT-6 architecture (left), where each transformer encoder comprises query, key, and value (QKV) attention, projection, and feedforward layers (middle). The QKV attention comprises 12 heads with 64 dimension output nodes per head and input token, resulting in $768 (= 12 \times 64)$ output dimensions.

## 3. Computational capability of a head using the confusion matrix

The functionality of each of the 12 heads in the six-transformer encoder block of the BERT-6 architecture pre-trained on a small Wikipedia dataset comprising 90,000 paragraphs [17] was quantified using the following method. The first five transformer encoder blocks and QKV attention of the six-transformer encoder block remained unchanged (frozen). The $128 \times 768$ output nodes of the scaled dot-product attention were connected to a classifier head trained on 90,000 Wikipedia paragraphs to minimize the loss function (Fig. 2, left). This small training dataset was selected to overcome our limited computational resources and was shown to maintain the qualitative results obtained using the entire Wikipedia dataset [17]. The accuracy of this partial BERT-6 architecture was estimated using 90,000 validation dataset comprising 28,273 tokens. The results indicate an average accuracy per token (APT) of 0.36 for ~23,188 tokens that were predicted correctly at least once in the masked process.

    The functionality of each of the 12 heads was estimated by silencing the 768 input nodes to the trained classifier head, except for the 64 that belonged to the examined head

(as exemplified for a pair of nodes in Fig. 2, right). Consequently, the 30,522 output units representing the tokens were influenced only by the 64 nodes of that head. The validation dataset was then propagated through the first five pre-trained transformer blocks, as well as through the unsilenced output units of the QKV attention of the sixth block, generating a 30,522 × 30,522 confusion matrix. Each matrix element $(i, j)$ represents the number of times an output unit $j$, representing token $j$, was selected by validation inputs with masked token $i$ (Fig. 2, right).

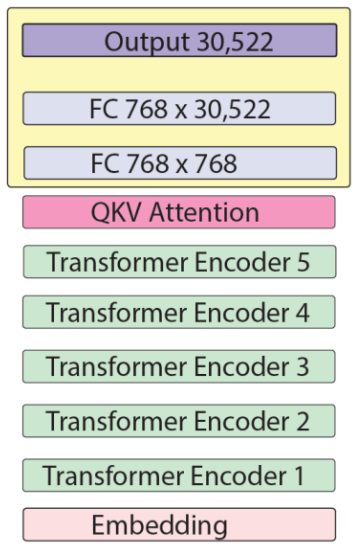
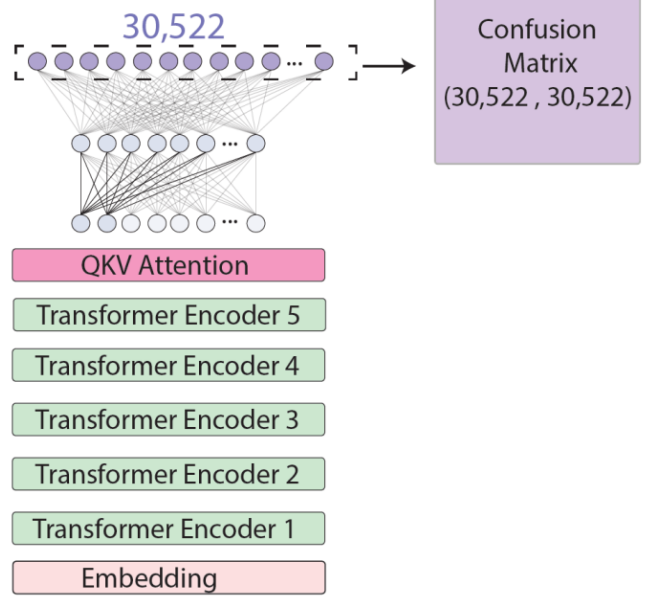

**Fig. 2.** A classifier head (yellow) connected to the 768 QKV attention output units of the 6[th] transformer encoder is trained using 90,000 Wikipedia paragraphs to minimize the loss function (left). All weights of the first fully connected layer of the classifier head are silenced (light gray) except for the emerging weights from a pair of selected nodes (black). The 30,522 × 30,522 confusion matrix is generated using a validation dataset, which is used to quantify a subset of nodal performance.

The number of positive diagonal elements of the confusion matrix, representing mask tokens that were correctly predicted at least once, varied among the 12 heads with average of 3,507 (Fig. 3A). The APT of the positive diagonal elements (0.043) fluctuates slightly among heads (Fig. 3B); this value is three orders of magnitude greater than the APT value of a random-guess (1/30,522). The diagonal confidence fluctuates slightly among heads, yielding an average of 0.176 (Fig. 3C), and is defined as the ratio between the sum of the positive diagonal elements of the confusion matrix and their corresponding columns. This measure represents the ratio of the total number of correctly predicted mask tokens to the total number of predicted outcomes for these tokens. The fraction of elements in the occupied columns with zero diagonal elements is less than a few percent; hence, the diagonal confidence and the confidence obtained using the full confusion matrix are typically very similar.

The computational capability of each of the heads is evident (Fig. 3), where the total number of tokens recognized by all of the heads was 13,243 out of the 28,273 in the validation dataset. Notably, most of the unrecognized tokens were those with low frequencies. For frequencies greater than 100 the heads recognized 8,428 tokens out of 9,386 in the dataset. The focusing of each head on a subset of tokens represents spontaneous symmetry breaking, as there is no directed order in the learning process to assign a specific subset of tokens for each head. This occurred following the random initial conditions of the weights and the biases of each one of the heads. Notably, in case of deterministic dynamics, that is, no dropouts [25] or stochastic gradient descent [26, 27], starting from the same initial conditions for the 12 heads within each layer, their weights will remain the same throughout the learning process. Consequently, there is one head per layer, and the learning almost vanishes.

The APT of each individual head (Table 1) was considerably lower than the APT obtained for the entire twelve heads ($\sim$0.36), indicating the importance of cooperation among the heads. The effect of cooperation among $n$ heads is estimated, similarly to the single-head procedure (Fig. 2, middle), by silencing the 768 input nodes to the trained classifier head except the $n \cdot 64$ nodes that belong to the examined heads. The results indicate an increase in the average APT with $n$, simultaneously with the number of positive diagonal elements and diagonal confidence (Table 1). The training process splits the learning task among the heads using spontaneous symmetry breaking, which is represented by the confusion matrices; however, there are additional hidden correlations among the output fields of the heads, contributing further to the average APT. This gap was attributed to events in which a mask token was selected following the summation of the output fields emerging from the 12 heads, contradicting the decision following the individual confusion matrices. Hence, cooperation among the output fields of the 12 heads enhanced the APT.

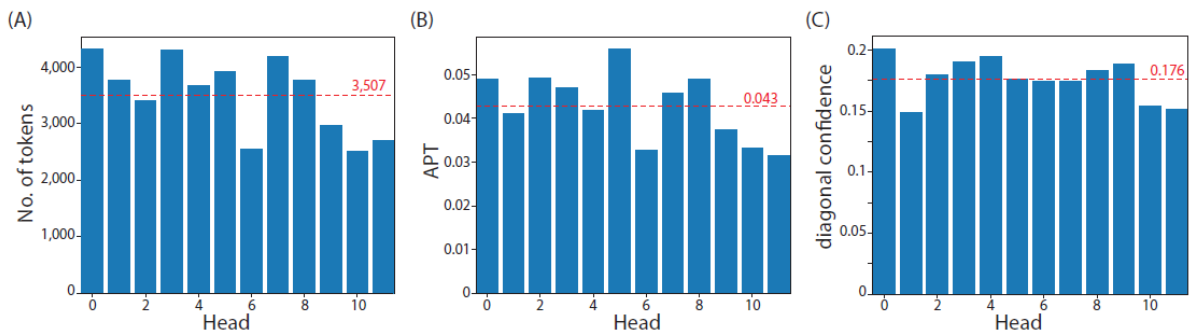

**Fig. 3.** Performance of the twelve heads in the six-transformer encoder. (A) Number of tokens with positive diagonal elements in each one of the confusion matrices. (B) Average APT per head for tokens with positive diagonal elements. (C) Diagonal confidence measuring the ratio between the sum of the diagonal positive elements of each confusion matrix and their sum of corresponding columns. The average for each panel is denoted (dashed horizontal red-line).

| No.of Heads | No. of tokens | APT | Diagonal confidence |
|---|---|---|---|
| 1 | 3,507 | 0.043 | 0.176 |
| 2 | 9,867 | 0.093 | 0.286 |
| 4 | 18,061 | 0.209 | 0.455 |
| 12 | 23,188 | 0.365 | 0.614 |

**Table 1.** Properties of the confusion matrix combining the output of several heads (first column). Number of tokens with positive diagonal elements (second column), and their average APT and diagonal confidence values (third and fourth columns) as a function of the number of heads. The results of each row denote the average over the twelve heads.

## 4. Spontaneous symmetry breaking at the nodal level

The computational capability of a single node, as well as a subgroup of nodes in the six-encoder block of the pre-trained BERT-6, was evaluated similarly to the estimation of the functionality of a single head. The first five transformer encoder blocks and the QKV attention of the six-transformer encoder block remained unchanged (frozen), and their $128 \times 768$ output units were connected to a classifier head trained to minimize the loss function (Fig. 2, left). The functionality of a subset of the QKV output nodes was estimated by silencing the $768$ input nodes to the trained classifier head, except for the selected nodal subset (Fig. 2, middle). Consequently, the $30,522$ output units representing the tokens were influenced solely by the limited aperture of a small subgroup of nodes or a single node. The results were based on an analysis of the confusion matrix generated using a limited aperture.

The confusion matrix number of positive diagonal elements increases from $\sim 3.7$ for one node to $\sim 3500$ for $64$ nodes (Fig. 4A and Table 1). The results were almost insensitive to whether the subgroup of nodes was selected from one head or at random from the $768$ inputs. A random output guess among the average of $\sim 3.7$ possible selected input tokens results in an APT of $1/3.7 \sim 0.27$. The obtained average APT value of $0.405$ (Fig. 4B) indicates an enhancement in the learning capability of a single node compared with the APT value of a random-guess. Nevertheless, the diagonal confidence is very low $\sim 0.03$, as almost all the $\sim 28,000$ mask tokens in the validation set are predicted as the output of the $3.7$ diagonal tokens.

In the case of two input nodes, there are on average $8.1$ possible selected output tokens and a measured diagonal APT value of $0.233$ (Fig. 4A), which is considerably enhanced compared with a random-guess APT value of $1/8.1 \sim 0.123$. As the number of input nodes is increased up to $12$, the number of classified tokens increased, whereas the average diagonal APT decreased. In this region, the decrease in the random-guess APT, $1/(No. of\ classified\ tokens)$, dominates the enhancement by cooperative learning using a few input nodes, resulting in a decrease in the average diagonal APT (Fig. 4B). However, a crossover occurred when the number of input nodes exceeds 12 (Figs. 4B and 4D), where the average diagonal APT increased as a function of the number of input nodes, although the random-guess APT decreased. This crossover indicates enhanced cooperative learning among the input nodes compared with the decreased baseline following a random-guess. In contrast to the non-monotonicity of the average diagonal APT (Figs. 4B and 4D), the diagonal confidence was monotonic. Almost all the mask tokens are predicted as the output of one of the diagonal tokens, and the probability of finding a non-zero column with zero diagonal element is very low. The number and sum values of the positive diagonal elements increase as a function of the number of input nodes, whereas the sum of their corresponding columns decreases, resulting in monotonically increased diagonal confidence.

The phenomenon where high APT values are obtained for a small number of tokens and a zero APT value for the rest (Fig. 4A) indicates spontaneous symmetry breaking by a few nodes, and even by a single node. In the training procedure, there was no preference for assigning a small subset of tokens to a given node or a small subset of nodes. This spontaneous symmetry breaking occurs following the initial conditions of the network and probably also following the selected order of the training inputs.

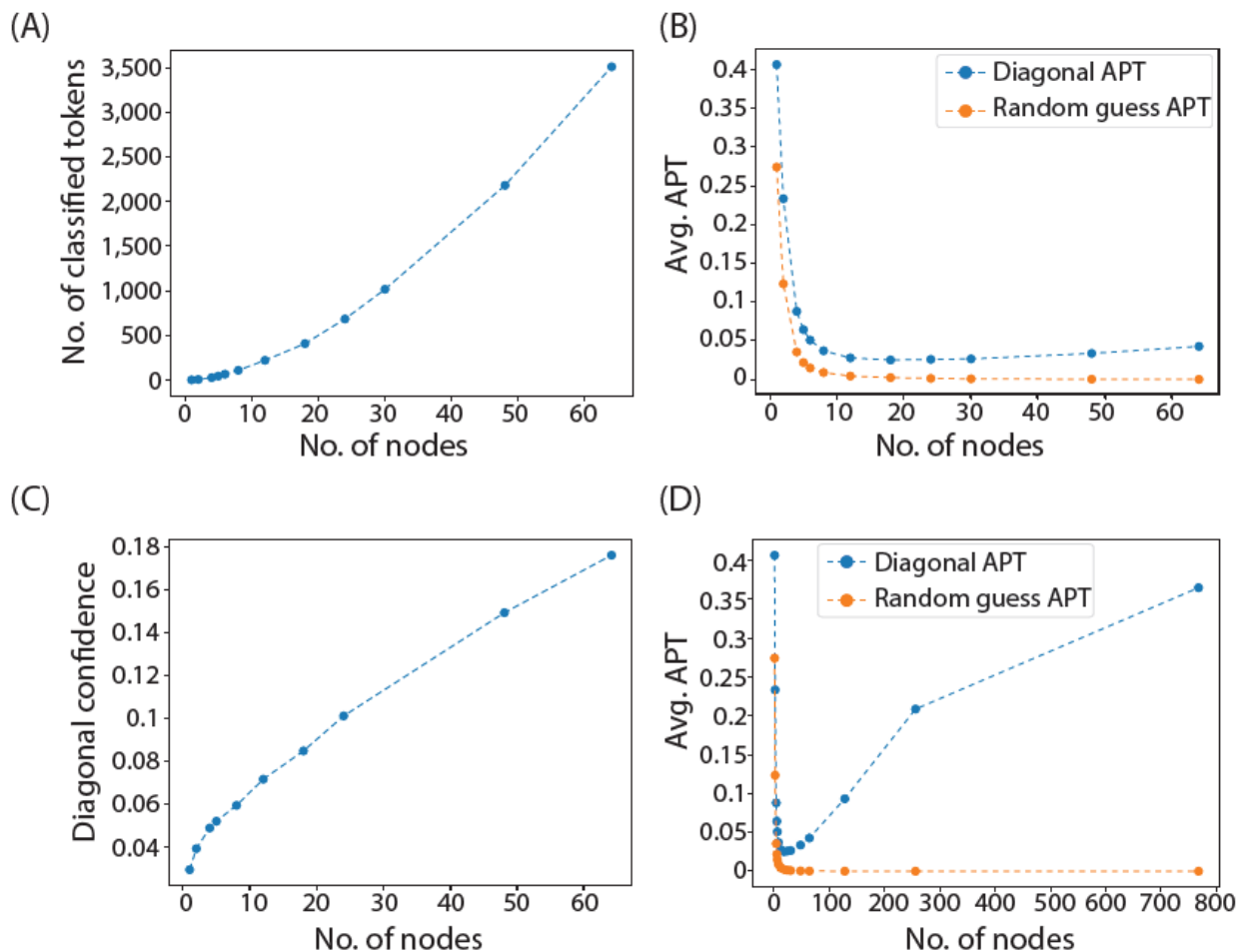

**Fig. 4.** Pre-training performance as a function of the number of output attention nodes of the six-transformer encoder. (A) Number of classified tokens selected as the positive diagonal elements in the confusion matrix. (B) Average accuracy per token (APT) of the diagonal tokens (blue) and average APT of a random-guess among the diagonal tokens (orange). (C) Diagonal confidence measuring the ratio between the sum of positive diagonal elements and their columns. (D) The same description as that listed in panel B applies for up to 768 output attention nodes.

## 5. FewRel classification using a few nodes

A similar phenomenon of spontaneous symmetry breaking by a small subset of nodes as observed in the pre-training process also occurs in a fully supervised FewRel classification task [27, 28], comprising 64 output labels, each with 630 training and 70 test instances.

A pre-trained BERT-6 over 90,000 small Wikipedia dataset was subsequently fine-tuned for the FewRel classification task. Subsequently, the 768 output nodes of the attention of the six-transformer encoder block were trained using a classifier head comprising only one $768 \times 64$ FC layer. The functionality of a selected subset of nodes was estimated by silencing 768 input nodes to the trained classifier head, excluding the selected subset of nodes (Fig. 2, middle) [11]. Consequently, the 64 output units representing the FewRel labels are influenced solely by the limited aperture of a small subgroup of nodes or a single node [8]. The results were based on the analysis of the $64 \times 64$ confusion matrix generated by such a limited aperture.

The number of positive diagonal elements in the confusion matrix increased from ~4.5 for a single node to ~62 for 64 nodes (Fig. 5A). For one node, the average diagonal accuracy of ~0.36 exceeded the random-guess accuracy of $1/4.5 \sim 0.22$, indicating an efficient classification even with a single attention output node (Fig. 5B). For the two input nodes, there are ~9 positive diagonal elements, and an average diagonal accuracy of 0.23 is considerably higher than the random-guess accuracy of $1/9 \sim 0.11$. Compared with the average diagonal accuracy, which is non-monotonic, the diagonal confidence increases monotonically as a function of the number of input nodes, whereas almost all inputs predict one of the diagonal labels (Fig. 5C, similar to 4C).

When the number of input nodes increased up to six, the number of classified labels, positive diagonal elements in the confusion matrix, increased, whereas the average diagonal accuracy decreased (Figs. 5A and 5B). In this region, the decrease in accuracy was dominated by the random-guess accuracy, $1/(No. of\ classified\ labels)$, over the enhancement by cooperative learning using a few input nodes (Fig. 5B). A crossover occurred when the number of input nodes exceeds six (Figs. 5B and 5D), similar to the observed crossover in the diagonal APT value (Fig. 4B).

The results clearly indicate spontaneous symmetry breaking in a classification task by a few nodes, and even by a single node, without any preference in the training procedure, similar to the phenomenon observed in the pre-training procedure (Fig. 4).

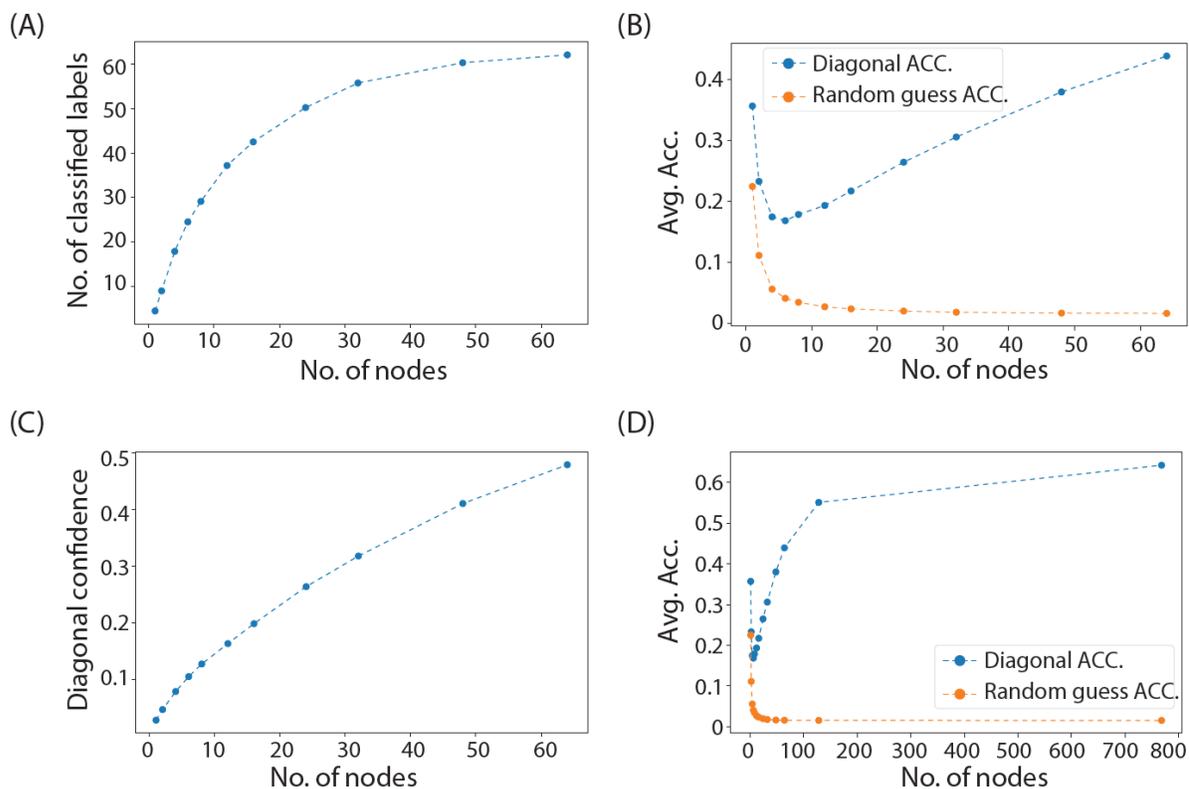

**Fig. 5.** Classification performance by a subset of output attention nodes of the six-transformer encoder using the FewRel dataset. (A) Number of classified labels - positive diagonal elements in the $64 \times 64$ confusion matrix. (B) Average accuracy (Acc.) of the diagonal labels (blue) and the average Acc. following a random-guess among the diagonal labels (orange). (C) Diagonal confidence. (D) The same description as that listed in panel B applies for up to $768$ output attention nodes.

## 6. Upper bound for classification using a few nodes

The computational capability of a few attention output nodes was sufficient to predict several mask tokens or classify FewRel labels with high accuracy, with performance improving as the number of output nodes increased. Nevertheless, this computational capability efficiency is in question because, theoretically, even a single node connected to the 30,522 output units with certain weights and biases can distinguish between all possible tokens for selected input ranges. Nevertheless, the objective of learning is to generate an appropriate input range common for all the input instances belonging to the same class, resulting in a correctly selected output. Hence, quantifying the achievable computational efficiency of a few nodes given their weights and biases requires a mathematical approach. This approach calculates the upper bound for distinguishable classes by assuming all the possible input ranges that are not necessarily realized by the learning process. Analyzing the gap between distinguishable classes realizable by the network with the mathematical approach can shed light on the role and the efficiency of spontaneous symmetry breaking with respect to the learning task.

For a fine-tuned BERT-6 using the fully supervised FewRel dataset, the upper bound for the computational capability of a few nodes can be analyzed using a trained classifier head comprising one layer, from the $768$ outputs of the six-transformer encoder to the $64$ output nodes. Given the fine-tuned weights emerging from a few input nodes to the $64$ output units and their biases, the maximum achievable number of positive diagonal confusion matrix elements can be quantified, assuming unconstrained input ranges. This upper bound can be obtained using a standard result from a convex hull analysis. The output space dimension is defined as the number of input nodes plus one, such that for each output node, a point is assigned according to its incoming weights and bias. A point can only maximize a linear function if it lies on the supporting face of the convex hull [29]. The convex hull is the smallest convex set that contains all points (intuitively, the "rubber-band" envelope around them). Only the upper part of the hull is retained, i.e., faces that are "visible from above" along the bias axis [30]. The unique vertices appearing on these upward-facing faces correspond exactly to the number of output nodes that can be maximized for a subset of nodes for unconstrained inputs. Counting these output nodes yielded an upper bound for the number of positive diagonal elements in the confusion matrix.

Applying this type of convex hull analysis while using the fine-tuning weights and biases of the one-layer classifier head indicates that for one input node, the number of positive diagonal elements in the confusion matrix, recognized labels, is very close to the upper bound (Fig. 6A). The gap between the number of recognized labels and the upper bound increased in the cases of two and four input nodes (Figs. 6B and 6C); while the variance among different subsets of nodes was small. The increasing gap as a function of the number of input nodes suggests that the learning process is non-optimal. It is likely that some additional labels recognized by the upper bound were excluded because of the limited classifier head input ranges utilized by the trained network.

The upper bound for the recognized labels of the classifier head was similar to the bound achieved by both the randomly initialized and fine-tuned BERT-6. Hence, the upper bound does not evolve considerably during the learning process. Nevertheless, the main difference between a random system in comparison to a trained system is the number of predicted labels and non-zero columns in the confusion matrix, which is approximately two, independent of the size of the input nodes. This observation indicates that learning implemented by spontaneous symmetry breaking mainly directs a class of input instances to a limited specific input range and selects the correct label.

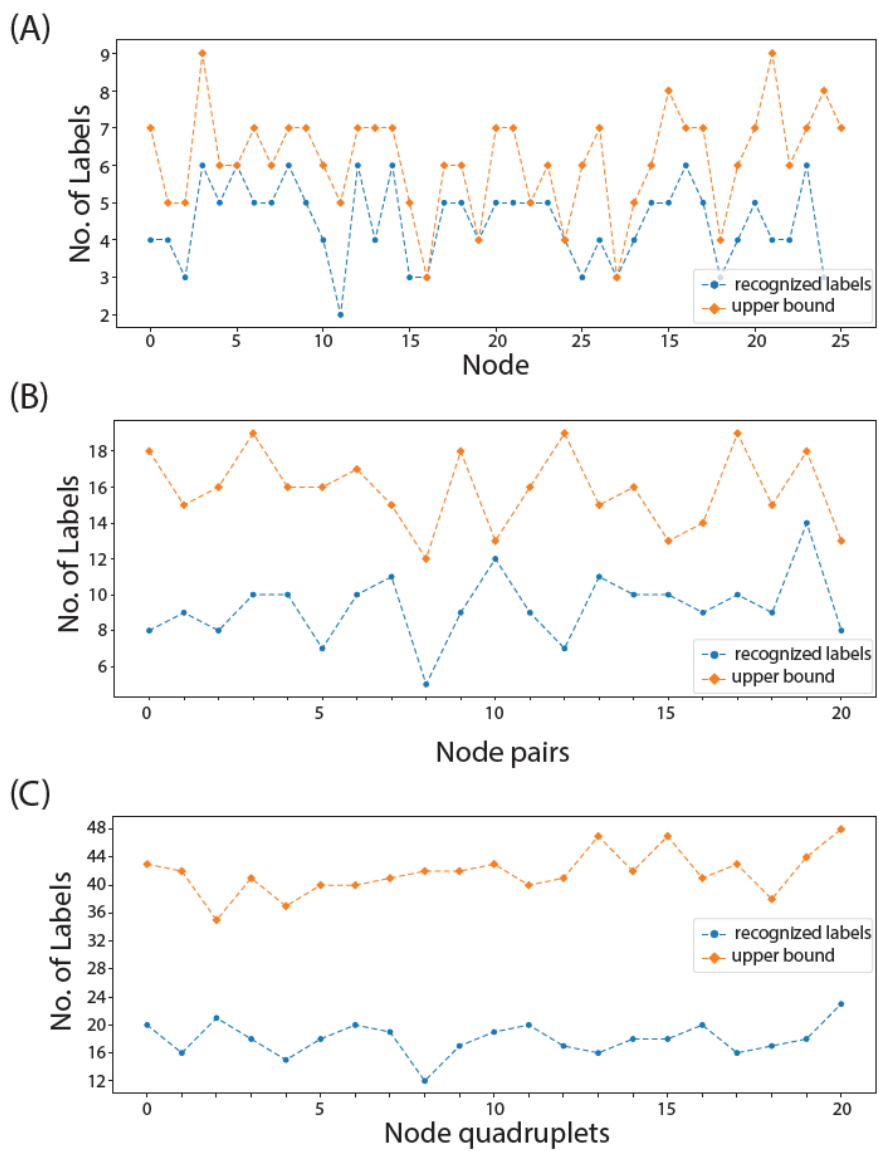

**Fig. 6.** Number of recognized labels using a small subset of input nodes of a trained classifier head to the 64 output units of the fine-tuned BERT-6 on the FewRel dataset. (A) Number of positive diagonal elements of the confusion matrix, i.e., the number of recognized labels for 25 single input nodes (blue) and for the upper bound for the same nodes (orange). (B) The description is the same as that listed in panel A but for 20 pairs of input labels. (C) The description is the same as that listed in panel A but for 20 sets of quadruplets input nodes.

## 7. Discussion

Spontaneous symmetry breaking in statistical mechanics primarily occurs in phase transitions wherein the Hamiltonian obeys, for instance, inversion symmetry, and the low-temperature free energy has a reduced symmetry. The emergence of this phenomenon typically requires two conditions: stochastic dynamics at finite temperatures and ergodic time divergence at the thermodynamic limit — infinite systems. The mechanism that prefers one of the symmetries of the system can be a non-extensive external field at the boundary of the bulk material, particularly in the case of second-order phase transitions, an infinitesimal random field on each degree of freedom, or initial conditions of the degrees of freedom with a small bias to one of the system's symmetries. The presented results indicate the emergence of spontaneous symmetry breaking for NLP tasks, even under zero-temperature dynamics, deterministic pre-training and fine-tuning procedures, such as those without dropout [25] and with a finite training architecture.

The spontaneous symmetry breaking at the level of a single and a small subset of nodes differs from a similar phenomenon in the phase transition of disordered systems, such as spin glasses. For instance, in an Ising spin-glass without an external magnetic field, the Hamiltonian obeys inversion symmetry, where below the transition temperature a frozen spin breaks its inversion symmetry [31-34]. However, by observing the frozen state of a single spin or a small subset of spins, one cannot deduce the equilibrium state of the entire system, that is, the frozen directions of all the spins. The goal of a physical system is to minimize the free energy which its macroscopic state cannot be deduced from the frozen information of a microscopic number of spins. In contrast to spin-glass systems, the examined NLP models indicate that the spontaneous symmetry breaking of a small number of nodes is related to the implementation of the network's task.

The phenomenon of APT crossover as a function of the number of examined input nodes was also found to be relevant to the FewRel accuracy that may indicate a universal behavior that governs other tasks and architectures. This is supported by the following argument adopted from the analysis of the binary classification capacity of a committee machine, which represents a majority decision among $K$ Perceptrons [35, 36]. For $K = 1$ and random binary inputs, the capacity per weight is $2$ [37, 38], whereas for $K = 3$, the capacity per weight is $\sim 3$ [39] which increased further at increasing $K$ values [40]. For example, the increased capacity for $K = 3$ beyond 2 is attributed to the fact that the output is determined by two, out of the three, Perceptrons which correctly predict the desired output, where there are three such legal internal representations. Hence, an input–output relation must be learned by cooperation between two Perceptrons only, which enhances the capacity per weight beyond 2. The number of legal internal representations grows exponentially as a function of the number of inputs, and the learning task is to maximize the capacity by assigning similar internal representations for correlated inputs. The degeneracy of the legal internal representations increases exponentially according to $2^{K-1}$, resulting in increasing capacity.

In the examined NLP tasks, binary inputs and outputs were replaced by continuous fields. The constraint for correct classification requires the strongest output field for the correct output label. This constraint can be generalized by the fact that the maximal field on the correct output label is the result of the sum of the output fields attributed to several input nodes. As the number of input nodes increases, the number of recognized labels increases, and the following two conflicting trends coexist. First, a linear decrease in the accuracy is expected following a random decision. Second, the possibility of enhanced learning following field summation emerging from several input nodes increases exponentially, similar to the increased number of legal internal representations with $K$ units in the committee machine. For a small number of input nodes, the first trend dominates, and the accuracy decreases with an increase in the number of recognized labels, whereas for a larger number of input units, the second trend dominates, and a crossover occurs.

Finally, the reported phenomenon of spontaneous symmetry breaking among the heads—as measured by the number of positive diagonal elements in the confusion matrix—becomes more prominent along the transformer blocks. It can be used to optimize the learning architecture and to enhance the learning algorithm. From a biological point of view, learning using a single node agrees with neuronal plasticity in the form of dendritic learning enabling a single neuron to acquire powerful computational capability [41-44]. Nevertheless, the presented results are based on a single pre-trained architecture using a small training dataset and one classification task. They require validation in other NLP tasks and datasets using enhanced computational resources.

# Appendix

*1. Dataset and preprocessing*

The datasets used in this study are FewRel [18] and Wikipedia [45]. Each dataset was tokenized using the BERT-base-uncased tokenizer from the HuggingFace Transformers library [16, 46], which converts raw text into 30,522 token IDs. Tokenization was performed using the following configuration: truncation to a maximum length of 128 tokens or padding to the same length.

*2. Optimization*

The CrossEntropyLoss [47] function was selected for the classification task and minimized using the stochastic gradient descent algorithm [26, 48] and the AdamW optimizer [49] was used. The maximal accuracy was determined by searching through the hyper-parameters (see below). The L2 regularization method [50] was applied during training.

To pre-train our models, we employed a Masked Language Modeling (MLM) objective similar to that used in the original BERT architecture [16]. Following standard masking strategy [46], 15% of tokens were selected for prediction. Of these, 80% were replaced with [MASK] token, 10% with random tokens, and the remaining 10% were left unchanged. Special tokens (e.g., [PAD], [CLS], [SEP]) were excluded from masking.

*3. Hyper-parameters*

The hyper-parameters $\eta$ (learning rate) and $\alpha$ (L2 regularization) were optimized for offline learning, using a mini-batch size of $32$ inputs. The learning-rate decay schedule was also optimized. For all the simulations, a linear scheduler was applied using the HuggingFace utility [46], with zero warm-up steps and a total number of training steps equal to the number of epochs. This schedule gradually causes the learning rate to decay from its initial value to zero in a linear fashion throughout training, which helps stabilize convergence [51].

The pre-training models in all the simulations were trained for 50 epochs with $\eta = 5.5e-5, \alpha = 1e-2$ and a linear scheduler was applied [17]. The pre-training of the classifier head for the Wikipedia dataset was trained for 50 epochs with $\eta = 1e-4, \alpha = 8e-6$ and a linear scheduler was applied. The same setup was examined for the pre-trained model without dropout, which yielded comparable spontaneous symmetry breaking for APT and classification on the level of a head as well as for a small subset of attention output nodes and a single output node, where the average APT and classification were only slightly reduced.

The fine-tuning on the FewRel task was applied for the pre-trained model on $90,000$ randomly selected subsets of Wikipedia [45] dataset [17] which was trained for $50$ epochs with $\eta = 5.5e-5, \alpha = 1e-2$ and a linear scheduler was applied. The fine-tuning on the classifier head for the FewRel dataset was trained for $20$ epochs with $\eta = 5e-5, \alpha = 8e-6$ and a linear scheduler was applied.

For Figs. 1-4, pre-training was performed on $90,000$ randomly selected subsets of Wikipedia [45] dataset. The classifier head didn't include biases and was trained on the same training dataset. Silenced nodes were applied only on the input to the classifier head.

For Fig. 5, the pre-training was performed on $90,000$ randomly selected subsets of Wikipedia dataset [45]. The classifier head comprised of one FC layer with biases and was fine-tuned on FewRel training dataset. As before, silencing was applied only at the input to the classifier head.

*4. Confusion matrix*

The confusion matrix was constructed using an MLM-style evaluation procedure. Instead of randomly masking tokens, we deterministically masked each non-special token in every sentence (one token at a time) across $90,000$ validation samples. As a result, each token was evaluated exactly as many times as it appeared in the dataset (i.e., according to its empirical frequency), thereby eliminating run-to-run variability across different testing iterations.

From the resulting confusion matrix, we compute the Accuracy Per Token (APT) by dividing each diagonal element by the sum of its corresponding row (considering only tokens with nonzero frequency).

*2.4.* Statistics

Statistics for all results were obtained using several samples and the standard division was around 1% for all the results.

2.5. Hardware and software

We used Google Colab and its available GPUs. We used Pytorch for all the programming processes.